\documentclass{article} 
\usepackage[preprint]{colm2026_conference}
\definecolor{linkblue}{RGB}{0,0,139}      
\definecolor{navy}{RGB}{0,0,128}          
\definecolor{royalblue}{RGB}{65,105,225}  
\definecolor{steelblue}{RGB}{70,130,180}  
\definecolor{dodgerblue}{RGB}{30,144,255}
\definecolor{mediumblue}{RGB}{0,0,205}    
\definecolor{darkgreen}{rgb}{0.0, 0.5, 0.0}
\definecolor{darkslateblue}{RGB}{72,61,139}
\definecolor{mydarkgreen}{rgb}{0,0.6,0}
\definecolor{table-blue}{RGB}{173, 216, 230}
\definecolor{mydarkgreen}{rgb}{0,0.6,0}
\usepackage{microtype}
\usepackage{url}
\usepackage[colorlinks = true,
            linkcolor = navy,
            urlcolor  = navy,
            citecolor = navy,
            anchorcolor = navy]{hyperref}
\usepackage[misc]{ifsym}
\usepackage{minitoc}
\usepackage{dsfont}
\usepackage{subcaption}
\usepackage{cancel}
\usepackage{booktabs}
\usepackage{colortbl}
\usepackage{amsmath}
\usepackage{mathtools}
\usepackage{amssymb}
\usepackage{graphicx}   
\usepackage{booktabs}   
\usepackage{algorithmic} 
\usepackage{algorithm}
\usepackage{verbatim}   
\usepackage{makecell}   
\usepackage{array}      
\usepackage{multirow}
\usepackage{xurl}
\usepackage{amsthm}
\usepackage{booktabs}
\usepackage{adjustbox}
\usepackage{wrapfig}
\usepackage{caption}
\usepackage{tabularx}
\newtheorem{definition}{Definition}
\usepackage[most,skins,theorems]{tcolorbox}
\usepackage{lineno}

\definecolor{darkblue}{rgb}{0, 0, 0.5}
\hypersetup{colorlinks=true, citecolor=darkblue, linkcolor=darkblue, urlcolor=darkblue}

\title{From $P(y|x)$ to $P(y)$: Investigating Reinforcement Learning in Pre-train Space}

\author{Yuqiao Tan$^{1,2}$\thanks{Equal contribution}, Minzheng Wang$^{1,2}$\footnotemark[1], Bo Liu$^{3}$, Zichen Liu$^{3}$, Tian Liang$^{4}$, Shizhu He$^{1,2}$\thanks{Corresponding author}, \\
\textbf{Jun Zhao$^{1,2}$, Kang Liu$^{1,2}$}  \\
$^1$ Institute of Automation, Chinese Academy of Sciences \\
$^2$ University of Chinese Academy of Sciences\\
$^3$ National University of Singapore
$^4$ Tencent AI Lab \\
\texttt{\{tanyuqiao2025,wangminzheng2023\}@ia.ac.cn, \{shizhu.he\}@nlpr.ia.ac.cn}
}

\begin{document}

\ifcolmsubmission
\linenumbers
\fi

\maketitle

\begin{abstract}
    While reinforcement learning with verifiable rewards (RLVR) significantly enhances LLM reasoning by optimizing the conditional distribution $P(y|x)$, its potential is fundamentally bounded by the base model's existing output distribution. Optimizing the marginal distribution $P(y)$ in the \textbf{Pre-train Space} addresses this bottleneck by encoding reasoning ability and preserving broad exploration capacity. Yet, conventional pre-training relies on static corpora for passive learning, leading to a distribution shift that hinders targeted reasoning enhancement. In this paper, we introduce \textbf{PreRL (Pre-train Space RL)}, which applies reward-driven online updates directly to $P(y)$. We theoretically and empirically validate the strong gradient alignment between $\log P(y)$ and $\log P(y|x)$, establishing PreRL as a viable surrogate for standard RL. Furthermore, we uncover a critical mechanism: \textbf{Negative Sample Reinforcement (NSR)} within PreRL serves as an exceptionally effective driver for reasoning. NSR-PreRL rapidly prunes incorrect reasoning spaces while stimulating endogenous reflective behaviors, increasing transition and reflection thoughts by 14.89$\times$ and 6.54$\times$, respectively. Leveraging these insights, we propose \textbf{Dual Space RL (DSRL)}, a Policy Reincarnation strategy that initializes models with NSR-PreRL to expand the reasoning horizon before transitioning to standard RL for fine-grained optimization. Extensive experiments demonstrate that DSRL consistently outperforms strong baselines, proving that pre-train space pruning effectively steers the policy toward a refined correct reasoning subspace. Code is available \href{https://github.com/Trae1ounG/Pretrain_Space_RLVR}{here}.
\end{abstract}

\begin{figure}[!h]
    \centering
    \includegraphics[width=1\linewidth]{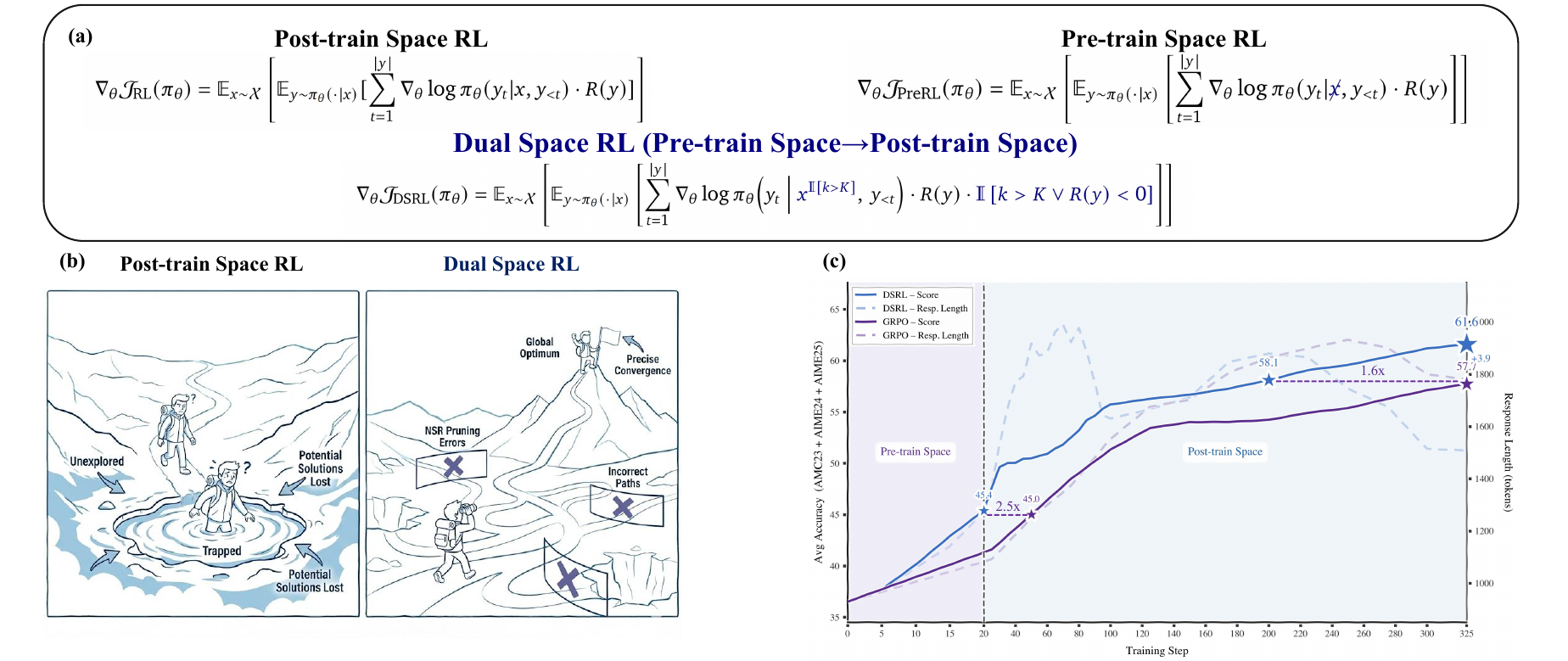}
    \caption{
    (a) Gradient objectives of Post-train Space RL, Pre-train Space RL and their combination in Dual Space RL.  (b) Conceptual comparison between standard RL (trapped in local optima) and DSRL (broad exploration via NSR pruning toward a refined correct reasoning subspace) (c) Training dynamics of DSRL vs. GRPO, where DSRL achieves higher average accuracy with improved sample efficiency and sustained response length growth driven by NSR-PreRL's exploration.
}
    \label{fig:intro}
\end{figure}

\section{Introduction}

Reinforcement learning (RL) has emerged as a key driver in advancing the complex reasoning capabilities of large language models (LLMs)~\citep{sutton1998reinforcement,gpt4,gemini-thinking,yang2025qwen3}. Notably, the success of DeepSeek-R1 has solidified reinforcement learning with verifiable rewards (RLVR) as a potent post-training paradigm for enhancing LLMs across diverse domains~\citep{guo2025deepseek,team2026kimi}.

Existing work has demonstrated that incentivized reasoning paths already exist within the base model's output distribution, suggesting that the reasoning capacity of RLVR-trained models may be fundamentally bounded by the capabilities of the base model~\citep{yuedoes, zuofar, peng2025simko,chen2025pass}. \textbf{Pre-train Space} optimization focuses on strengthening this foundation by directly targeting $P(y)$~\citep{gupta2023continual,huang2026remit}. The core value of this paradigm lies in its deep and broad knowledge internalization: by encoding reasoning capability directly into model parameters, it establishes an enhanced foundation that preserves broad exploratory capacity, and provides stronger initialization for subsequent optimization of $P(y|x)$ within the \textbf{Post-train Space}.

However, this paradigm has a fundamental limitation: pre-training relies on static corpora for passive learning, leading to a pronounced distribution shift between pre-trained knowledge and the task distribution encountered during post-training, which hinders targeted reasoning enhancement~\citep{zhoumegamath,pasteropenwebmath,zhang2025interplay,wang2025octothinker}. In contrast, RL approaches have achieved remarkable capability gains by shifting the output distribution toward higher-reward trajectories, thereby improving sampling efficiency during online optimization.

This observation motivates a natural idea: \textbf{introducing RL into the pre-train space to synergize knowledge internalization with reward-driven optimization}. 
Built upon this stronger initialization, post-train RL then focuses on fine-grained optimization of $P(y|x)$, concentrating on conditional trajectory refinement and leading to better exploration.

In this paper, we explore the feasibility of \textbf{Pre-train Space RL (PreRL)}. As shown in Figure~\ref{fig:intro} (a), we optimize the marginal distribution $P(y)$, which removes the question condition during updates. We provide theoretical justification and empirical verification that the gradients of $\log P(y)$ and $\log P(y|x)$ are strongly aligned, which ensures that optimizing the marginal distribution effectively improves the conditional policy, offering a viable surrogate that naturally preserves the exploration capacity and encourages  high rewards trajectories.

Dissecting the roles of positive and negative samples in PreRL~\citep{zhu2025surprising} reveals a critical asymmetry. Specifically, Positive Sample Reinforcement (PSR) in the pre-train space degrades performance by accumulating probability mass on self-generated correct samples, whereas Negative Sample Reinforcement (NSR) proves to be an exceptionally instructive mechanism.
We find that NSR-PreRL rapidly prunes incorrect trajectories in the pre-train space, promoting enhanced exploration in the post-train space. Simultaneously, it stimulates the model's endogenous reasoning capabilities, with transition and reflection thoughts increasing by $14.89\times$ and $6.54\times$ respectively, effectively broadening the search space through richer probabilistic exploration~\citep{chen2025seal, zeng2025simplerl}.

Leveraging these insights, we adopt the \textbf{Policy Reincarnation} strategy~\citep{liang2025squeeze, agarwal2022reincarnating} to synergize NSR-PreRL with standard RL which learns through pre-train space to post-train space, called \textbf{Dual Space RL (DSRL)}. 
In this framework, the model is initialized via NSR-PreRL to expand its reasoning horizon and elicit intrinsic reasoning capabilities, and subsequently switched to standard RL for direct policy optimization. Extensive experiments demonstrate that DSRL consistently outperforms strong baselines, achieving superior accuracy with better sample efficiency and robust Pass@$K$ improvements.

\begin{itemize}
   \item We introduce \textbf{Pre-train Space RL (PreRL)}, a novel paradigm that optimizes the marginal distribution P(y) within RL. We theoretically and empirically validate the alignment between $P(y)$ and $P(y|x)$, establishing PreRL as a viable surrogate for standard RL that naturally preserves the model's intrinsic exploration capacity.
    \item We identify Negative Sample Reinforcement (NSR) in the pre-train space as an exceptionally inspirational mechanism. NSR-PreRL effectively prunes incorrect reasoning paths while stimulating endogenous reasoning behaviors, and achieving comparable accuracy with fewer training steps than standard RL.
    \item We adopt the Policy Reincarnation strategy to synergize NSR-PreRL with standard RL into \textbf{Dual Space RL (DSRL)}. DSRL consistently outperforms standard RL and achieves higher Pass@$K$, indicating that pre-train space learning steers the policy toward a refined correct reasoning subspace.
\end{itemize}

\section{From Post-train Space $P(y|x)$ to Pre-train Space $P(y)$}

We first formally define the optimization spaces for LLMs. In the context of RL for LLMs, the policy $\pi_\theta$ is equivalent to the output distribution $P$ of LLMs parameterized by $\theta$. Notably, the fundamental distinction between these spaces lies in how the policy is \textit{optimized}.

\begin{definition}[\textbf{Post-train Space Optimization}] 
Given an input distribution $\mathcal{X}$, the paradigm of Post-train Space optimization is defined as the optimizing the conditional policy $\pi_\theta(y|x)$, where the learning objective strictly conditions the update on the specific input query $x \sim \mathcal{X}$. 
\end{definition} 

\begin{definition}[\textbf{Pre-train Space Optimization}] 
The Pre-train Space optimization is defined as optimizing the marginal policy $\pi_\theta(y)$, which directly optimizes the intrinsic distribution of reasoning trajectories. We provide a detailed comparison with traditional pre-training in Appendix~\ref{app:comparison}.
\end{definition}

We visualize the token probability distributions over the vocabulary to analyze their relationship. The conditional and marginal distributions often exhibit similar token rankings, representing aligned cases (Figure~\ref{fig:token_distribution}), though significant misalignments also occur (Figure~\ref{fig:token_distribution_app}). As detailed in Figure~\ref{fig:prob_comparison}, log-probabilities align closely for high-probability, deterministic tokens, but diverge notably for early-sequence or highly uncertain tokens.

\subsection{Post-train Space Reinforcement Learning: $P(y|x)$}
Language model generation can be formulated as a token-level Markov Decision Process (MDP). At each step $t$, the state $s_t = [x;y_{<t}]$ consists of the input question and generated tokens so far. The policy $\pi_\theta(\cdot | s_t)$ samples the next token $y_t$ from vocabulary, transitioning to $s_{t+1}=[s_t;y_t]$. Generation terminates upon producing \texttt{[eos]} or reaching the budget. 
To optimize the policy in the post-train space, standard RL maximizes the expected return conditioned on inputs:
\begin{equation}
\begin{aligned}
    \label{eq:rl_objective}
    \mathcal{J}_\text{RL}(\pi_\theta) =
    \underset{{x \sim \mathcal{X}}}{\mathbb{E}}\left[ \underset{y \sim \pi_\theta(\cdot|x)}{\mathbb{E}}[R(y)]  - \beta \mathbb{D}_{KL}[\pi_\theta(\cdot|x)) || \pi_{\text{ref}}(\cdot|x)]\right],
\end{aligned}
\end{equation}
where $R(y)=\sum_{t=1}^{|y|}r(y_t)$ is the cumulative return, $|y|$ is the length of $y$ and $\pi_{\text{ref}}$ is the reference policy. We adopt sparse rewards where $r_t = 0$ for $t < |y|$ and $r_{|y|} \in \{0, 1\}$ indicates task success.
Assuming $\beta=0$ for simplicity following~\citet{hu2025openreasonerzeroopensourceapproach}, the gradient of the objective function is given by the Policy Gradient theorem~\citep{sutton1998reinforcement}:
\begin{equation}
    \nabla_\theta \mathcal{J}_\text{RL}(\pi_\theta)=\mathbb{E}_{x\sim\mathcal{X}}\left[\mathbb{E}_{y\sim\pi_\theta(\cdot|x)}[\sum_{t=1}^{|y|}\nabla_\theta\log\pi_\theta(y_t|x,y_{<t})\cdot R(y)]\right].
\end{equation}
\textbf{Policy Optimization.} We employ GRPO~\citep{shao2024deepseekmath} as our primary optimization algorithm with the optimized token-level policy gradient loss from~\citet{yu2025dapo}. For each input query $x$, GRPO samples a group of $G$ responses $\{y_1, \dots, y_G\}$ from the old policy $\pi_{\theta_\text{old}}$ and computes their corresponding returns $\mathbf{R} = \{R_1, \dots, R_G\}$. The advantage for each token in response $y_i$ is then estimated by normalizing the returns within the group: $\hat{A}_{i,t}=\frac{R_i - \operatorname{mean}(\mathbf{R})}{\operatorname{std}(\mathbf{R})}$. The optimization objective is formulated as:
\begin{equation} 
\label{eq:grpo} 
    \mathcal{J}_{\text{GRPO}}(\pi_\theta) =   
    \mathbb{E}_{x\sim \mathcal{X},\{y_i\}^G_{i=1}\sim\pi_{\theta_\text{old}}} 
    \left[ \frac{1}{\sum^G_{i=1}{|y_i|}}\sum^G_{i=1}\sum_{t=1}^{|y_i|}\min \left(\rho_{i,t}\hat{A}_{i,t}, \text{clip}(\rho_{i,t}, 1-\epsilon, 1+\epsilon)\hat{A}_{i,t}\right) \right], 
\end{equation} 
where $\rho_{i,t}=\frac{\pi_\theta(y_{i,t}|x, y_{i,<t})}{\pi_{\theta_\text{old}}(y_{i,t}|x, y_{i,<t})}$ denotes the importance sampling ratio, and $\epsilon$ is the clipping parameter to constrain policy updates. This group-relative advantage estimation effectively reduces variance and stabilizes training without a value network~\citep{schulman2015high}.

\subsection{Pre-train Space Reinforcement Learning: $P(y)$}
Pre-training paradigms operate by optimizing the entire sequences to encode knowledge while preserving an effective exploration space. Inspired by this, we propose \textbf{PreRL (Pre-train Space Reinforcement Learning)}, which directly optimizes the marginal policy $\pi_\theta(y)$ instead of the conditional policy $\pi_\theta(y|x)$. The policy gradient of PreRL is defined as:
\begin{equation} 
    \nabla_\theta \mathcal{J}_\text{PreRL}(\pi_\theta)=\mathbb{E}_{x\sim\mathcal{X}}\left[\mathbb{E}_{y\sim\pi_\theta(\cdot|x)}\left[\sum_{t=1}^{|y|}\nabla_\theta\log\pi_\theta(y_t|{\color{navy} \cancel{x}},y_{<t})\cdot R(y)\right]\right],
\end{equation} 
where ${\color{navy}\cancel{x}}$ signifies that PreRL removes the input dependency during the gradient update.

However, a critical question remains: \textit{\textbf{how does optimizing the marginal log-likelihood $\log\pi_\theta(y_t|y_{<t})$ affect the conditional log-likelihood $\log \pi_\theta(y_t|x,y_{<t})$ required for the task?}} 

\begin{figure}[!t]
    \centering
    \includegraphics[width=\textwidth]{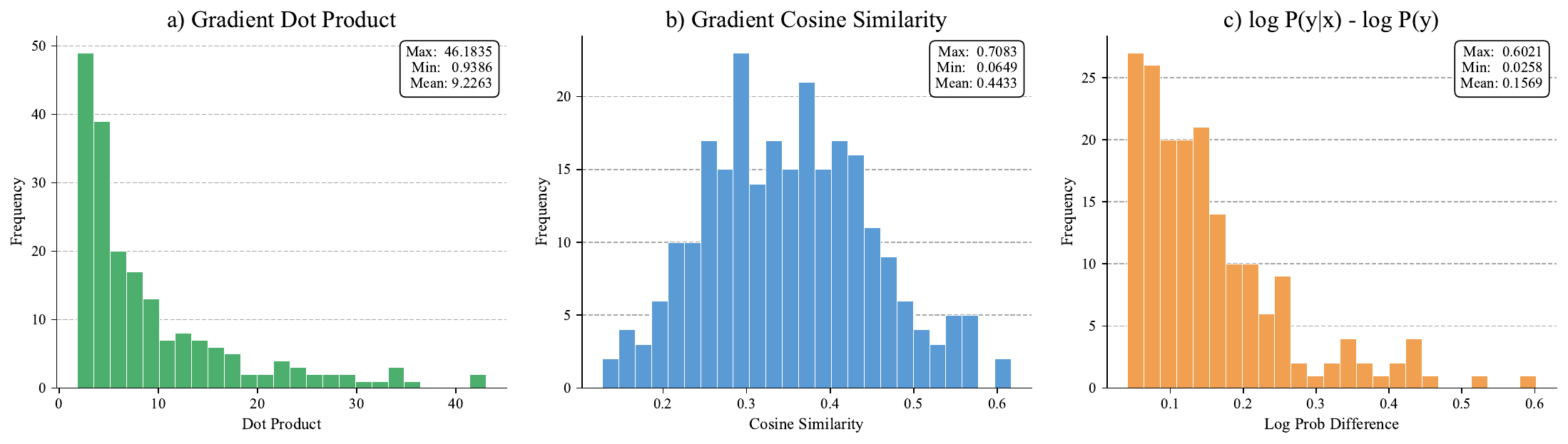}
    \caption{Synergistic effect analysis of $\log P(y|x)$ and $\log P(y)$ using Qwen3-4B on AMC23. (a) Gradient dot product between $ \nabla_\theta \log \pi_\theta(y)$ and $\nabla_\theta \log \pi_\theta(y|x)$. (b) Gradient cosine similarity. (c) Per-token log probability difference between $\log P(y|x)$ and $\log P(y)$.}
    \label{fig:synergy}
\end{figure}

Following \citet{hu2025test}, our theoretical justification is grounded in a fundamental premise regarding \textbf{Shared Parameter Influence}: the model parameters $\theta$ simultaneously govern both the marginal distribution $\pi_\theta(y)$ and the conditional distribution $\pi_\theta(y|x)$.

Building on this premise, we show that optimizing $\log \pi_\theta(y)$ improves $\log \pi_\theta(y|x)$ when their gradients align. Let $\theta' = \theta + \eta \nabla_\theta \log \pi_\theta(y) \cdot R(y)$ denote the parameters after one PreRL step updating the marginal log-probability. The first-order Taylor expansion yields:
\begin{align} 
        \log \pi_{\theta'}(y|x) \approx \;& \log \pi_\theta(y|x) + \eta  \cdot R(y) \cdot \underbrace{\left[ \nabla_\theta \log \pi_\theta(y) \right]^\top \nabla_\theta \log \pi_\theta(y|x)}_{\text{cross-gradient term}} + \mathcal{O}(\eta^2). 
\end{align}
A core premise underlying PreRL's effectiveness is that for reasoning trajectories $y$ which semantically aligned with $x$, the gradients of the marginal and conditional objectives exhibit a non-negative inner product: $\langle \nabla_\theta \log \pi_\theta(y), \nabla_\theta \log \pi_\theta(y|x) \rangle \geq 0$. This ensures the cross-gradient term is non-negative, meaning that updating the marginal  $\log \pi_\theta(y)$ concurrently influences the conditional  $\log \pi_{\theta}(y|x)$ in the same direction. Consequently, this alignment validates PreRL as an effective surrogate for standard RL. To empirically validate this, we conduct a detailed analysis on Qwen3-4B using 400 rollouts from the AMC23~\citep{AMC23}.

\textbf{Observation 1: The gradient inner product is consistently non-negative.}
We evaluate the synergy between the marginal and conditional objectives via $\langle \nabla_\theta \log \pi_\theta(y), \nabla_\theta \log \pi_\theta(y|x) \rangle$. As Figure~\ref{fig:synergy}(a) visualizes, the empirical distribution is entirely non-negative, satisfying the condition for 100\% of samples with an average inner product of $+9.2$. Figure~\ref{fig:synergy}(b) reinforces this with a strongly positive cosine similarity distribution. Ultimately, this high gradient alignment confirms that optimizing the $\pi_\theta(y)$ inherently benefits the $\pi_\theta(y|x)$.

\textbf{Observation 2: The distributions of $\log\pi_\theta(y|x)$ and $\log\pi_\theta(y)$ are closely aligned.}
As Figure~\ref{fig:synergy}(c) shows, the strongly overlapped distributions validate $\log\pi_\theta(y)$ as a faithful surrogate for $\log\pi_\theta(y|x)$. By operating in the pre-train space, this optimization internalizes question-agnostic reasoning, facilitating broad task transfer and enhanced general capacity.

 \begin{figure}[!t]
    \centering
    \includegraphics[width=\textwidth]{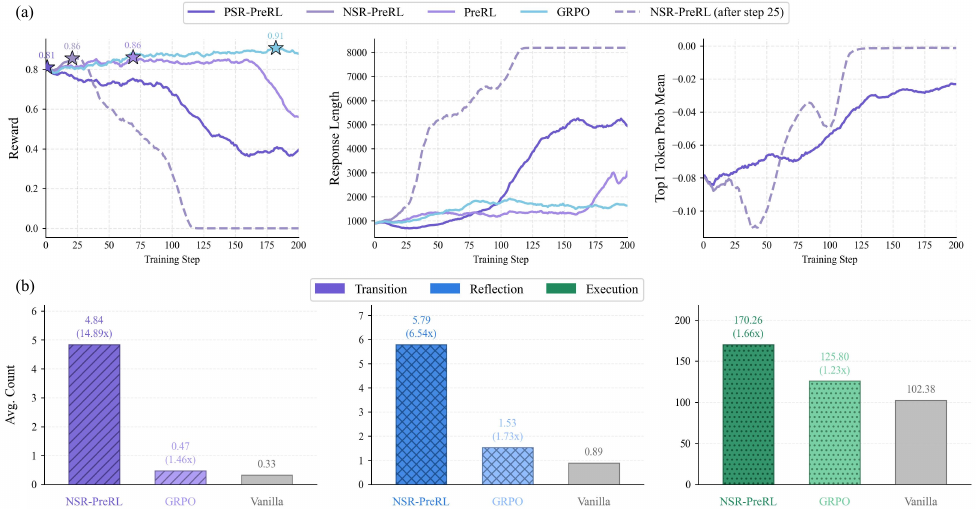}
    \caption{(a) Training dynamics of PreRL v.s. RL. Effects on reward, response length and top-1 token log-probability. (b) Comparison of distinct reasoning thoughts. NSR-PreRL trains for 20 steps and GRPO for 25 steps. The backbone model is Qwen3-4B.}
    \label{fig:prerl_dynamics}
\end{figure}

\subsection{In-depth Analysis of Pre-train Space Reinforcement Learning}
To understand PreRL's dynamics, we implement it using GRPO as the base algorithm. As Figure~\ref{fig:prerl_dynamics}(a) shows, PreRL achieves comparable performance to standard RL for the first 150 steps before suffering a significant collapse. To investigate this failure, we decompose the optimization into positive and negative sample reinforcement (PSR and NSR), following \citet{zhu2025surprising}. While PSR maximizes $\log \pi_\theta(y|x)$ to reinforce correct responses, NSR minimizes it to suppress incorrect ones. In PreRL, this dichotomy is governed by the advantage: samples with $A_i > 0$ act as positive signals and those with $A_i < 0$ as negative.

\textbf{PSR-PreRL suffers from on-policy learning collapse.} 
Unlike PSR-RL, PSR-PreRL targets $\log\pi_\theta(y)$. While their gradient directions align, their learning dynamics diverge. As Figure~\ref{fig:prerl_dynamics}(a) shows, PSR-PreRL successfully increases the conditional probability $\pi_\theta(y|x)$, corroborating their synergistic effect. However, it fails to effectively learn from self-generated on-policy trajectories, ultimately degrading performance. In contrast, QFFT~\citep{liuqfft} successfully optimizes the same objective $\max_\theta \pi_\theta(y)$ using out-of-distribution long-CoT trajectories from a teacher model. This suggests that maximizing $\pi_\theta(y)$ in the pre-train space strictly requires high-quality, out-of-distribution expert demonstrations, indicating a potential need for off-policy data integration during RL~\citep{yan2025learning}.

\textbf{NSR-PreRL alone demonstrates surprising effectiveness.} 
Negative samples contain undesired patterns and receive negative rewards to diminish their probability mass. As Figure~\ref{fig:prerl_dynamics}(a) shows, NSR-PreRL successfully decreases $\pi_\theta(y|x)$, supporting our prior analysis. Notably, NSR-PreRL strongly elicits reasoning capabilities, significantly increasing response length. We follow \citet{chen2025seal} to categorize reasoning steps into transition, reflection, and execution thoughts (Table~\ref{tab:criteria}). As shown in Figure~\ref{fig:prerl_dynamics}(b), after just 20 NSR-PreRL steps, the model generates $14.89\times$ more transitions and $6.54\times$ more reflections than the vanilla model, substantially outperforming GRPO. This confirms that NSR-PreRL activates internal knowledge for deeper reasoning. Consequently, this enhanced reasoning rapidly boosts performance, achieving an accuracy of 86\% with $3\times$ fewer training steps than standard RL.

However, this phenomenon is a double-edged sword. On one hand, it yields better performance and preserves exploration ability by redistributing probability mass away from incorrect trajectories in the pre-train space, effectively pruning wrong reasoning paths~\citep{zhu2025surprising}. On the other hand, it eventually leads to excessively long outputs that hinder continuous training~\citep{chen2024not, tan2025zero,wang2026mitigating}.

\begin{table*}[t!]  
    \centering
    \small
    \caption{Avg@32 results on MATH500, AMC23, AIME24, AIME25, Minerva and OlympiadBench. \textbf{Bold} and \underline{underlined} denote the best and second best.}
    \label{tab:main_exp}
    \resizebox{\textwidth}{!}{
    \begin{tabular}{lcccccccc}
        \toprule
        \textbf{Methods} & \textbf{AMC} & \textbf{MATH500} & \textbf{AIME24} & \textbf{AIME25} & \textbf{Minerva} & \textbf{OlympiadBench} & \textbf{Avg.} \\
        \midrule

        \multicolumn{8}{l}{\texttt{Qwen3-4B}} \\
        \quad Vanilla      & 68.28 & 80.17 & 23.13 & 20.00 & 23.62 & 32.33 & 41.26 \\
        \quad PPO          & 87.89 & \underline{89.64} & 47.50 & 41.15 & 30.19 & 40.52 & 56.15 \\
        \quad Reinforce++  & 79.37 & 86.88 & 35.83 & 30.00 & 27.30 & 38.15 & 49.59 \\
        \quad RLOO         & 85.00 & 89.38 & 45.10 & 40.62 & 30.27 & 40.46 & 55.14 \\
        \quad Dr.GRPO      & 87.58 & 89.22 & \underline{51.04} & \underline{42.60} & \textbf{31.05} & \underline{40.75} & \underline{57.04} \\
        \quad DAPO         & \underline{88.20} & 89.05 & 47.08 & 40.00 & 30.30 & 40.28 & 55.82\\
        \quad GRPO         & 87.81 & 89.17 & 46.46 & 40.94 & 30.06 & 40.29 & 55.79 \\
        \rowcolor{table-blue!50}
        \quad \textbf{DSRL (Ours)} & \textbf{89.22} & \textbf{89.68} & \textbf{51.15} & \textbf{43.44} & \underline{30.48} & \textbf{41.26} & \textbf{57.54} \\

        \midrule

        \multicolumn{8}{l}{\texttt{Qwen3-8B}} \\
        \quad Vanilla      & 69.53 & 80.24 & 25.73 & 19.27 & 22.33 & 32.61 & 41.62 \\
        \quad PPO          & \underline{89.69} & 89.40 & 51.98 & \underline{41.25} & \underline{30.30} & \underline{41.45} & \underline{57.34} \\
        \quad Reinforce++  & 84.14 & 88.61 & 47.40 & 35.10 & 29.19 & 39.52 & 53.99 \\
        \quad RLOO         & 87.27 & \underline{90.09} & 53.85 & 38.33 & 29.52 & 40.65 & 56.61 \\
        \quad Dr.GRPO      & 87.66 & 89.71 & 52.60 & 37.29 & 29.22 & 40.45 & 56.16 \\
        \quad DAPO         & 88.52 & 89.60 & 52.08 & 39.37 & 29.08 & 40.71 & 56.56\\
        \quad GRPO         & 88.05 & 89.91 & \underline{54.06} & 39.37 & 29.65 & 40.97 & 57.00 \\
        \rowcolor{table-blue!50}
        \quad \textbf{DSRL (Ours)} & \textbf{90.00} & \textbf{90.31} & \textbf{56.15} & \textbf{42.19} & \textbf{30.32} & \textbf{41.82} & \textbf{58.47} \\

        \bottomrule
    \end{tabular}
    }
\end{table*}

\section{Dual Space RL: Combining PreRL and RL via Policy Reincarnation} 
Our analysis reveals that NSR-PreRL excels by pruning incorrect reasoning trajectories in the pre-train space while eliciting deeper reasoning. To capitalize on these benefits, we adopt the \textbf{Policy Reincarnation} strategy~\citep{agarwal2022reincarnating, liang2025squeeze,tan2025bottom}: midway through training, we replace the base model with the NSR-PreRL checkpoint and restart on-policy RL. We term this sequential framework \textbf{Dual Space RL (DSRL)}, as it unifies the pre-train and post-train spaces. Its formulation is given by:

\begin{equation} 
    \nabla_\theta \mathcal{J}_{\text{DSRL}}(\pi_\theta) = 
    \mathbb{E}_{x \sim \mathcal{X}, y \sim \pi_\theta(\cdot|x)} \Big[ 
        \sum_{t=1}^{|y|} \nabla_\theta \log \pi_\theta(y_t | {\color{navy}x^{\mathds{I}[s > S]}}, y_{<t}) 
        \cdot R(y) \cdot {\color{navy}\mathds{I}[s > S \lor R(y) < 0]} 
    \Big], 
    \label{eq:dsrl} 
\end{equation} 
where $s$ denotes the current step, and $S$ specifies the transition threshold from pre-train space to post-train space optimization. The term $\color{navy}x^{\mathds{I}[s > S]}$ controls the conditioning: in PreRL phase when $s \le S$, the input $x$ is masked. Crucially, the indicator $\color{navy}\mathds{I}[s > S \lor R(y) < 0]$ enforces the NSR-PreRL constraint during the initial phase, ensuring that updates are only applied to negative samples to prune incorrect reasoning paths, while standard RL utilizes all samples in $s > S$. In Eq.~\ref{eq:dsrl}, we employ GRPO as the representative RL algorithm.

\section{Experiments}

\subsection{Experimental Setup}
\label{sec:exp_setup}
\textbf{Training Setup.} We employ Qwen3-4B and Qwen3-8B~\citep{yang2025qwen3} as base models and train on the MATH dataset~\citep{lewkowycz2022solving}, which contains 7,500 problems. Detailed training hyperparameters and setups are provided in Appendix~\ref{app:training_setup}.

\textbf{Evaluation Setup.} We evaluate DSRL against several RL baselines, including GRPO~\citep{shao2024deepseekmath}, PPO~\citep{schulman2017proximal}, Reinforce++~\citep{hu2025reinforce++} and RLOO~\citep{ahmadian2024back}. Also with the optimized version of GRPO: Dr.GRPO~\citep{liuunderstanding} and DAPO~\citep{yu2025dapo} with only clip higher mechanism. The benchmarks cover: MATH500~\citep{lightman2023let}, AMC23~\citep{AMC23}, AIME24~\citep{AIME24}, AIME25~\citep{AIME25}, Minerva~\citep{lewkowycz2022solving} and OlympiadBench~\citep{he2024olympiadbench}. 
Due to high output variance, we report Avg@32 and Pass@$K$ ($K \in [1, 256]$) estimated from $n=300$ samples per problem~\citep{chen2021evaluating}. We report more evaluation details in Appendix~\ref{app:training_setup}.

\subsection{Main Results}
\label{sec:main_results}
As shown in Table~\ref{tab:main_exp}, DSRL consistently improves Avg@$K$ over strong RL baselines across all benchmarks and models. For Qwen3-4B, DSRL significantly outperforms GRPO by 4.69 points on AIME24 and 2.50 points on AIME25. It also surpasses advanced methods like Dr.GRPO and DAPO by 0.50 and 1.72 points on average, respectively. Similarly, Qwen3-8B demonstrates robust gains, achieving a best average score of 58.47. Overall, these results indicate that initializing with NSR-PreRL internalizes crucial knowledge to build a stronger foundation, thereby enabling subsequent RL to guide the policy toward higher accuracy.

\begin{figure}[!t]
    \centering
    \includegraphics[width=\textwidth]{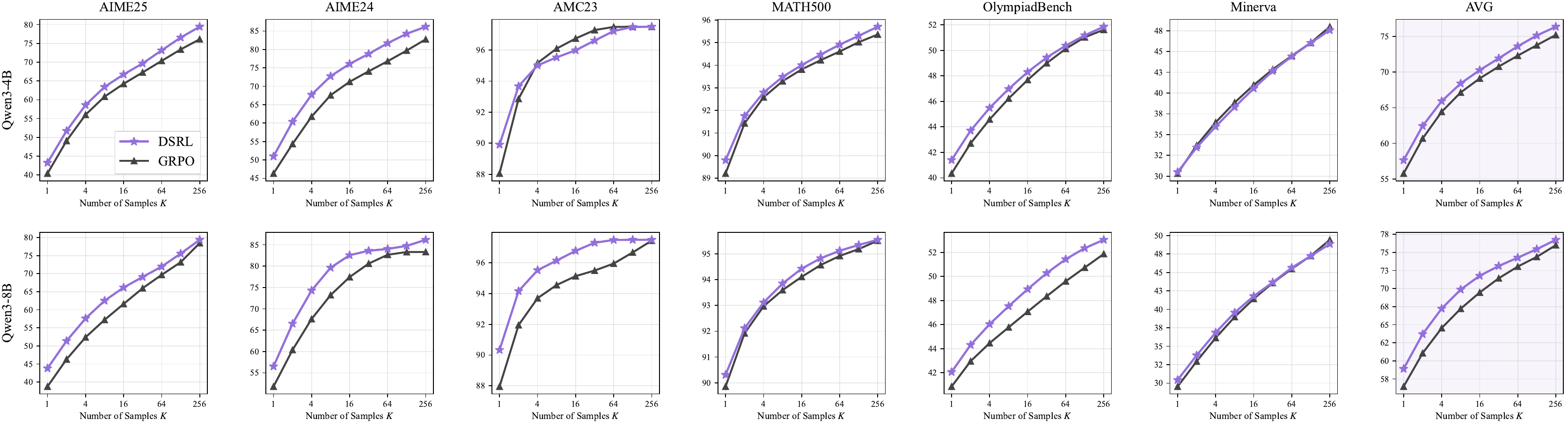}
    \caption{Pass@$K$ performance comparison between DSRL and GRPO across LLMs.}
    \label{fig:passk}
\end{figure}

\begin{figure}[!t]
    \centering
    \begin{minipage}[t]{0.44\textwidth}
    \centering
        \includegraphics[width=\textwidth]{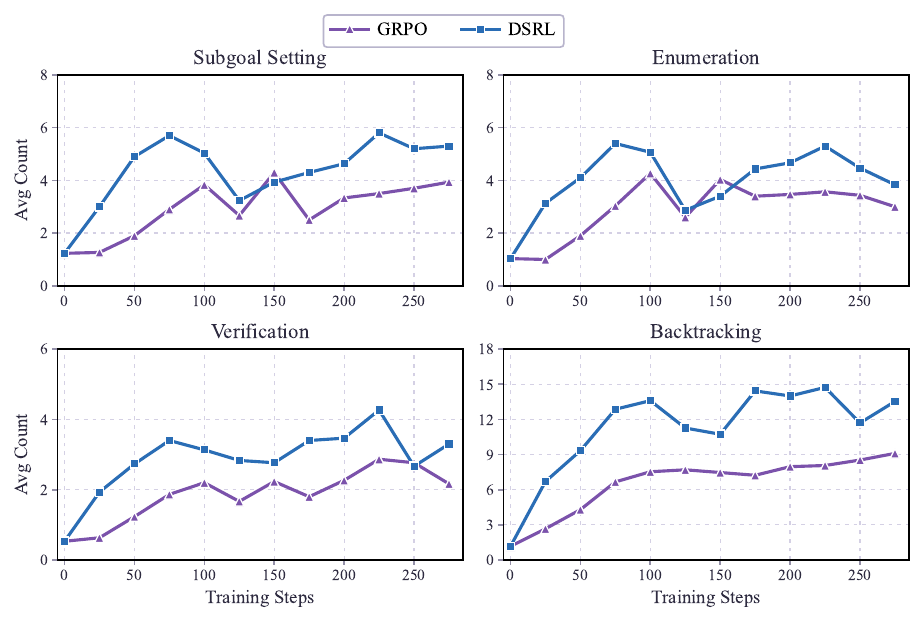}
        \caption{Reasoning behaviors over RL training with different methods.}
        \label{fig:reasoning_behavior}
        
    \end{minipage}
    \hfill
    \begin{minipage}[t]{0.53\textwidth}
        \centering
        \includegraphics[width=\textwidth]{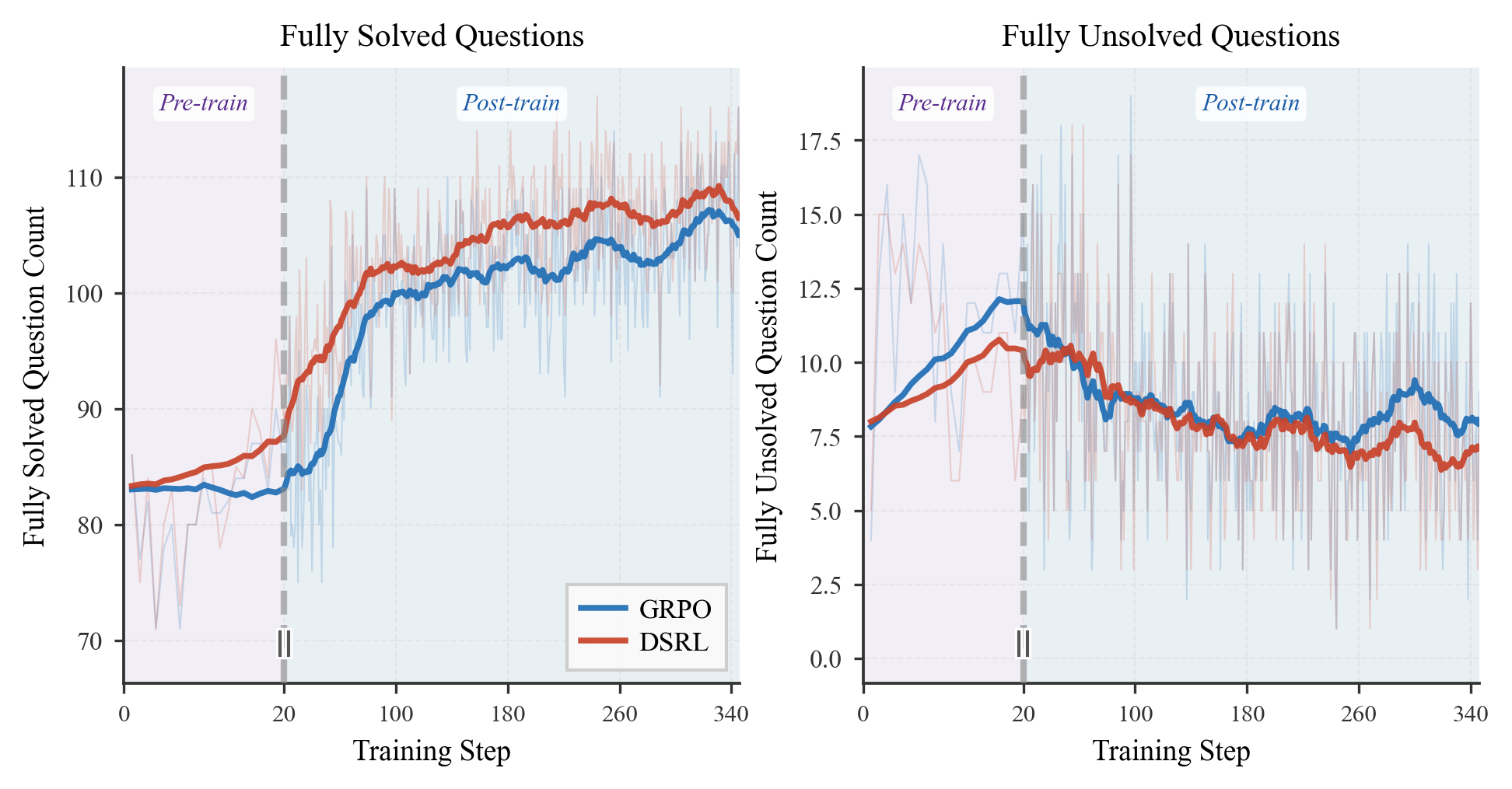}
        \caption{Evolution of problem-solving status (Solved vs.\ Unsolved) on training dataset.}
        \label{fig:behavior_count}
    \end{minipage}
\end{figure}
To comprehensively evaluate DSRL, we compare its Pass@$K$ performance against GRPO (Figure~\ref{fig:passk}). DSRL consistently outperforms the baseline across the entire range of $K$. On Qwen3-8B, DSRL achieves superior results for all $K$, maintaining a clear lead on challenging benchmarks like AIME24 and AIME25 as $K$ increases. This robust sampling scalability indicates that DSRL not only boosts Pass@1 but also diversifies the high-quality solution space. Similarly, Qwen3-4B results show DSRL outperforming the baseline across most $K$ budgets, efficiently concentrating probability mass on correct reasoning paths. Ultimately, these findings demonstrate that DSRL's pre-train space optimization fosters a more exploration-friendly policy landscape, fundamentally enhancing reasoning capacity.

\begin{figure}[!t] 
    \begin{minipage}[c]{0.55\textwidth}
        \centering
        \scriptsize
        \captionof{table}{OOD generalization results. \textbf{Bold} denotes the best.}
        \label{tab:ood_results}
        \setlength{\tabcolsep}{4pt}
        \begin{tabular}{lcccc}
        \toprule
        \textbf{Methods} & \textbf{GPQA-Diamond} & \textbf{MMLU-Pro} & \textbf{BBH} & \textbf{HumanEval}\\
        \midrule
        \multicolumn{3}{l}{\texttt{Qwen3-4B}} \\
        Vanilla  & 42.12 & 58.61 & 75.55 & 78.66 \\
        GRPO     & 39.39 & 61.12 & 80.37 & \textbf{81.10}\\
        \rowcolor{table-blue!50}
        \textbf{DSRL (Ours)} & \textbf{43.18} & \textbf{66.49} & \textbf{82.41} & \textbf{81.10} \\
        \rowcolor{table-blue!50}
        $\Delta$ vs.\ GRPO & \textcolor{blue}{+3.79} & \textcolor{blue}{+5.37} & \textcolor{blue}{+2.04} & +0.00 \\
        \midrule
        \multicolumn{3}{l}{\texttt{Qwen3-8B}} \\
        Vanilla  & 45.86 & 59.72 & 79.57 & 87.20\\
        GRPO     & 50.96 & 59.47 & \textbf{82.35} & 87.80\\
        \rowcolor{table-blue!50}
        \textbf{DSRL (Ours)} & \textbf{53.48} & \textbf{63.79} & 82.31 & \textbf{90.24}\\
        \rowcolor{table-blue!50}
        $\Delta$ vs.\ GRPO & \textcolor{blue}{+2.52} & \textcolor{blue}{+4.32} & -0.04 & \textcolor{blue}{+2.44} \\
        \bottomrule
        \end{tabular}
        \end{minipage}
        \hfill
        \begin{minipage}[c]{0.40\textwidth}
        \centering
        \includegraphics[width=\linewidth]{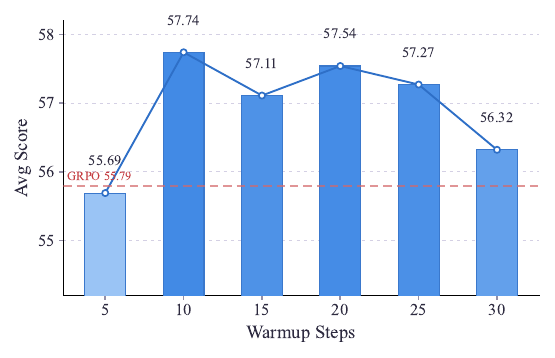}
        \captionof{figure}{Ablation on warmup steps. Bar color intensity reflects score magnitude.}
        \label{fig:warmup_ablation}
        \end{minipage}
    \end{figure}

To evaluate generalization beyond mathematical reasoning, we test on four out-of-distribution (OOD) benchmarks: GPQA-Diamond~\citep{rein2024gpqa}, MMLU-Pro~\citep{wang2024mmlu}, BBH~\citep{suzgun2023challenging} and HumanEval~\citep{chen2021evaluating}. As Table~\ref{tab:ood_results} shows, DSRL consistently matches and outperforms GRPO across both scales. Specifically, DSRL achieves substantial gains on knowledge-intensive tasks, improving GPQA-Diamond by +3.79 and MMLU-Pro by +5.37 for Qwen3-4B, with corresponding gains of +2.52 and +4.32 for Qwen3-8B. DSRL also generalizes well to code generation, achieving a +2.44 improvement on HumanEval for Qwen3-8B. These results indicate that pre-train space optimization not only enhances in-domain reasoning but also cultivates a highly generalizable policy with superior OOD transferability.

\subsection{Analysis}
\label{sec:analysis}

\textbf{Training Efficiency in Dual-Space RL.}
Figure~\ref{fig:intro}(c) illustrates DSRL's superior training dynamics over GRPO across three dimensions. First, DSRL consistently leads in performance, achieving a 61.6 average accuracy vs. GRPO's 57.7. Second, it demonstrates exceptional efficiency by reaching 45.0\% and 58.0\% accuracy with $2.5\times$ and $1.6\times$ fewer steps, respectively. 

\textbf{Evolution of Reasoning Behaviors.} 
Following \citet{zeng2025simplerl}, we track four key reasoning behaviors on AIME25 (Subgoal Setting, Enumeration, Verification, Backtracking). In Figure~\ref{fig:reasoning_behavior}, RL struggles with slow and limited behavior emergence due to its constrained initial warmup. Conversely, DSRL drives rapid, sustained growth and achieves significantly higher frequency ceilings across all patterns. This confirms that pre-train space optimization effectively removes conditional constraints, unlocking intrinsic capacities for both rigorous self-correction and complex, structured reasoning exploration~\citep{chen2025seal}.

\textbf{Foundation Reasoning Enhanced via NSR-PreRL.} 
To assess foundational problem-solving abilities, we track the count of \textit{Fully Solved} (all rollouts correct) and \textit{Fully Unsolved} (all rollouts incorrect) questions during training. These metrics indicate whether the model has genuinely internalized core error patterns rather than succeeding by chance. In Figure~\ref{fig:behavior_count}, DSRL achieves rapid mastery during the NSR-PreRL phase, marked by a sharp surge in \textit{Fully Solved} counts and a decline in \textit{Fully Unsolved} ones. This proves that NSR-PreRL systematically eliminates failure modes by internalizing fundamental error logic.
This efficiency strongly validates our policy reincarnation strategy. By pruning universal incorrect patterns during NSR-PreRL, DSRL establishes a robust foundation. The subsequent RL phase can then focus on refining problem-specific nuances. Consequently, DSRL ultimately achieves more \textit{Fully Solved} questions and a lower error rate, confirming that pre-train space optimization elevates the policy's reasoning ceiling beyond RL limits.

\subsection{Ablation Study}
   
\paragraph{Impact of Warmup Steps.} 
We investigate the optimal duration for the NSR-PreRL warmup phase in Figure~\ref{fig:warmup_ablation}. The results exhibit an inverted-U trend, with 10 to 25 steps emerging as the optimal range. Conversely, fewer steps provide insufficient stimulation, while excessive warmup acts as a double-edged sword, causing over-exploratory generation that hinder subsequent fine-grained optimization in the post-train space. Thus, a carefully chosen transition point $S \in [10, 25]$ is crucial for the success of policy reincarnation.

\begin{wraptable}{r}{0.42\textwidth}
    \centering
    \footnotesize
    \caption{Ablation on warmup strategy.}
    \label{tab:ablation_nsr_warmup20}
    \setlength{\tabcolsep}{4pt}
    \begin{tabular}{lc}
    \toprule
    \textbf{Methods} & \textbf{Avg.} \\
    \midrule
    \multicolumn{2}{l}{\texttt{Qwen3-4B}} \\
    Vanilla & 41.26 \\
    GRPO & 55.79 \\
    NSR-RL Warmup & 54.38 \\
    \rowcolor{table-blue!50}
    \textbf{NSR-PreRL Warmup (DSRL)} & \textbf{57.54} \\
    \bottomrule
    \end{tabular}
\end{wraptable}
\paragraph{Pre-train Space vs. Post-train Space Warmup.} 
To investigate whether negative samples reinforcement in the post-train space can achieve similar benefits, we compare DSRL against an NSR-RL warmup baseline for both 20 steps. 
As shown in Table~\ref{tab:ablation_nsr_warmup20}, NSR-RL warmup paradoxically underperforms GRPO with an average score of 54.38 compared to 55.79, whereas DSRL achieves the highest score of 57.54. 
As shown in Figure~\ref{fig:ablation_nsr_dynamic}, by step 20, NSR-PreRL outperforms NSR-RL by an average of 6.6 points across the AMC23, AIME24, and AIME25, providing a vastly superior initialization for the subsequent RL phase.
 
\section{Related Works}
\subsection{Reinforcement learning with verifiable rewards}
Reinforcement learning with verifiable rewards (RLVR)~\citep{guo2025deepseek} has emerged as an effective post-training paradigm for enhancing LLMs reasoning
~\citep{shao2024deepseekmath, cai2026vi,zhao2025redone, liu2026automated, wang2026adaptive,fang2026proximity,fang2026allocate}\nocite{liao2026resadapt,wang2025comprehensive}.
Since RLVR has been argued to merely over-sharpen the sample distribution of the base model, potentially constraining the exploration space~\citep{yuedoes}, a complementary line of work investigates the balance between exploration and exploitation in the post-train space RL~\citep{li2025choice,wang2026anchored, yu2026unveiling,guo2025tree}.
While these studies offer insights into how reward signals shape policy behavior, their scope remains confined to the conditional distribution $P(y|x)$. Whether and how such reward-driven mechanisms can operate on the marginal distribution $P(y)$ remains  unexplored. 

\subsection{Pre-train Space Optimization}
Directly training LLMs on downstream tasks may suffer from distribution shift induced by novel data~\citep{cossu2024continual}. Continual pre-training addresses this by encoding broad knowledge into parameters, optimizing the marginal distribution $P(y)$ through offline training on static corpora, and thereby providing a stronger foundation for downstream tasks~\citep{sun2020ernie, cossu2024continual,ou2025llms,su2025scaling}. Such pre-train space optimization establishes general knowledge enhancement, preserving broad exploration capacity that is conducive to subsequent RL~\citep{huang2026remit, wang2024mathpile, wang2025octothinker,zhoumegamath}.
Our work departs from them by bringing online RL directly into the pre-train space, transforming knowledge internalization from a passive, data-driven process into an active, reward-guided one. We also provide a detailed comparison between PreRL and the emerging Reinforcement Learning Pre-Training (RLPT) paradigm in Appendix~\ref{app:comparison_rlpt}.

\section{Conclusion}
In this paper, we propose \textbf{Pre-train Space RL (PreRL)}, integrating reward-driven reinforcement learning into the pre-train space. We theoretically and empirically demonstrate a strong gradient alignment between the conditional objective $\log P(y|x)$ and the marginal objective $\log P(y)$. Dissecting the PreRL, we reveal that Negative Sample Reinforcement (NSR) acts as an exceptionally effective driver by selectively pruning incorrect reasoning paths and eliciting endogenous reasoning capabilities. Building on this, we introduce \textbf{Dual Space RL (DSRL)}, which uses a Policy Reincarnation strategy to synergize NSR-PreRL with standard RL. By initializing with NSR-PreRL to eliminate fundamental errors and expand the exploration horizon, DSRL consistently outperforms strong baselines across extensive benchmarks. Ultimately, our work proves that pre-train space optimization inherently establishes a more robust reasoning foundation for superior final performance.
\bibliography{colm2026_conference}
\bibliographystyle{colm2026_conference}

\clearpage
\appendix

\section{Detailed Comparison with Related Works}
\subsection{Comparison of PreRL with Pre-training and Continual Pre-training}
\label{app:comparison}

While Pre-training, Continual Pre-training, and PreRL all operate in the pre-train space by optimizing the marginal distribution $P(y)$, they differ fundamentally in their data sources, learning signals, and knowledge acquisition mechanisms. We summarize the key distinctions in Table~\ref{tab:comparison}.

\begin{table}[h]
\centering
\caption{Comparison of Pre-training, Continual Pre-training, and PreRL. All three paradigms optimize the marginal distribution $P(y)$ in the pre-train space, but differ in data sources, learning signals, and training paradigms.}
\label{tab:comparison}
\resizebox{\textwidth}{!}{
\begin{tabular}{lccc}
\toprule
\textbf{Dimension} & \textbf{Pre-training} & \textbf{Continual Pre-training} & \textbf{PreRL (Ours)} \\
\midrule
Optimization Target & $P(y)$ & $P(y)$ & $P(y)$ \\
Training Data & General web text & Task-specific corpora & Reasoning tasks \\
Data Format & Raw documents & Documents / QA pairs & Response-only trajectories \\
Learning Signal & Next-token prediction & Next-token prediction & Verifiable rewards \\
Learning Paradigm & Offline, passive & Offline, passive & Online, active \\
Task Alignment & Low & Moderate & High \\
\bottomrule
\end{tabular}
}
\end{table}

\textbf{Pre-training} establishes general language capabilities by learning from large-scale web corpora (e.g., Common Crawl, Wikipedia) through Next-Token Prediction (NTP)~\citep{brown2020language,radford2019language,raffel2020exploring}. While it builds broad linguistic knowledge, the resulting distribution $P(y)$ is not aligned with downstream reasoning tasks, leading to a significant distribution gap when transitioning to post-training.

\textbf{Continual Pre-training} narrows this gap by further training on task-specific corpora, which may include domain-relevant documents, general-purpose responses, or question-answer pairs~\citep{gupta2023continual,wang2024mathpile,zhoumegamath}. Although this approach injects targeted knowledge into $P(y)$, it still relies on static, pre-collected data and passive NTP, limiting its ability to adapt to the model's evolving capabilities during training.

\textbf{PreRL} fundamentally departs from both paradigms by replacing passive data consumption with active, reward-guided learning. Rather than training on external corpora, PreRL generates reasoning trajectories online through self-rollouts and uses verifiable rewards to selectively reinforce or suppress these trajectories. Crucially, PreRL updates only on the response $y$ with the input question $x$ removed, maintaining the pre-train space objective of optimizing $P(y)$. This design transforms knowledge internalization from a static, data-driven process into a dynamic, feedback-driven one, ensuring that the optimized $P(y)$ remains tightly aligned with the task distribution encountered during subsequent RL.

\subsection{Comparison with Reinforcement Pre-Training Paradigm}
\label{app:comparison_rlpt}

To overcome the substantial verification data-wall, Reinforcement Learning Pre-Training (RLPT) has recently emerged as a promising direction. Several methods have attempted to integrate RL objectives directly into the pre-training phase~\citep{hatamizadeh2025rlp, dong2025reinforcement,li2025reinforcement,xing2025pretrainzero}. While these RLPT methods share the conceptual goal of applying RL beyond the standard post-training phase to enhance foundational reasoning capabilities, our proposed PreRL fundamentally differs from the general RLPT paradigm in data sources, learning signals, and optimization spaces. We summarize the key distinctions in Table~\ref{tab:comparison_rlpt}.

\begin{table}[h]
\centering
\caption{Comparison of the Reinforcement Pre-Training (RLPT) paradigm and PreRL. PreRL distinguishes itself by operating on real reasoning tasks with truly verifiable rewards, strictly optimizing the marginal distribution $P(y)$ through active online exploration.}
\label{tab:comparison_rlpt}
\resizebox{0.8\textwidth}{!}{
\begin{tabular}{lcc}
\toprule
\textbf{Dimension} & \textbf{RLPT Paradigm} & \textbf{PreRL (Ours)} \\
\midrule
Optimization Target & $P(y|x)$ & $P(y)$ \\
Training Data & Pre-training corpora & Reasoning tasks \\
Data Format & Full Reasoning trajectories & Response-only trajectories \\
Learning Signal & Next-token prediction & Verifiable rewards \\
Learning Paradigm & Online, active & Online, active \\
Task Alignment & Moderate & High \\
\bottomrule
\end{tabular}
}
\end{table}

To overcome the substantial verification data-wall, Reinforcement Learning Pre-Training (RLPT) has recently emerged as a promising direction. Several methods have attempted to conduct RL directly on the pre-training corpus~\citep{hatamizadeh2025rlp, dong2025reinforcement,li2025reinforcement,xing2025pretrainzero}. While these RLPT methods can be regarded as a effective warmup to subsequent RL, our proposed PreRL fundamentally differs from the general RLPT paradigm in training data, learning signals, and optimization spaces and so on. We summarize the key distinctions in Table~\ref{tab:comparison_rlpt}.

\textbf{Reinforcement Pre-Training (RLPT)} operates primarily on static pre-training corpora. Since raw corpora lack objective verification signals, RLPT methods typically convert the Next-Token Prediction (NTP) task into a pseudo-reward. Specifically, the model generates self-rollouts as intermediate thinking, and the ultimate reward is determined by whether these rollouts successfully predict the subsequent ground-truth tokens. While this approach attempts to elicit reasoning, it is fundamentally constrained by the proxy nature of the NTP reward. Unlike mathematical reasoning tasks that offer strictly verifiable and definitive answers, next-token continuations are inherently open-ended with multiple valid possibilities. Consequently, relying on a single static continuation renders the optimization objective simplistic and lacking in rigorous logical verification.

\textbf{PreRL} fundamentally departs from the RLPT paradigm by operating directly within the standard RLVR framework on reasoning tasks. Instead of relying on static dataset continuations and NTP pseudo-rewards, PreRL utilizes objective task success  as the verifiable reward. Crucially, the fundamental distinction lies in the optimization space: we restrict the policy update strictly to the \textit{pre-train space} by masking the question during gradient updates. This allows PreRL to actively explore reasoning trajectories online and effectively learn through verifiable rewards. By doing so, PreRL answers the foundational question of how to enhance reasoning capabilities autonomously, without the confounding factors of supervised interventions or static data grounding.

\section{Detailed Experiment Settings}
\label{app:experiment_settings}

\subsection{Training and Evaluation Setup}
\label{app:training_setup}

\textbf{Training Setup.} We employ Qwen3-4B and Qwen3-8B~\citep{yang2025qwen3} as the base models for all experiments, since they exhibit strong capability in advanced reasoning and use non-thinking mode~\citep{ma2025reasoning}.  
For the training set, we use MATH~\citep{lewkowycz2022solving}, which contains 7,500 problems. We train the models using the verl framework~\citep{sheng2025hybridflow}. The prompt batch size is 128, with 8 rollouts generated per prompt. The sampling temperature during training is set to 1.0, and the maximum response length
is set to 8192 tokens. We update the model with a mini-batch size of 128 and a learning rate of 1e-6 to conduct on-policy RL for 6 epochs~\citep{andrychowicz2020matters}. 
We sample a group of responses per question and estimate advantages using the group-based normalization.

\begin{table}[h]
\centering
\caption{RL Hyperparameters}
\label{tab:rl_hyperparameters}
\begin{tabular}{ll}
\toprule
\textbf{Hyperparameter} & \textbf{Value} \\
\midrule
Optimizer & AdamW \\
Policy learning rate & $1\text{e}^{-6}$ \\
Training batch size & 128 \\
Samples per prompt & 8 \\
Mini-batch size & 128 \\
Training epochs & 6 \\
Max prompt length & 1024 tokens \\
Max response length & 8192 tokens \\
Rollout temperature & 1.0 \\
\bottomrule
\end{tabular}
\end{table}

\textbf{Evaluation Setup.} For evaluation, We use vLLM~\citep{kwon2023efficient} with temperature=1.0 and top\_p=1.0.  Due to high output variance in reasoning tasks, we report Avg@32 (Pass@1 averaged over 32 rollouts). For Pass@$K$, which is defined as Pass@$K:=\mathbb{E}_{x\sim\mathcal{D}}\left[1-\binom{n-c}{K}/\binom{n}{K}\right]$,  where $c$ denotes the number of correct completions out of $n$ generated responses~\citep{chen2021evaluating}. To reduce evaluation variance on those datasets, we set $n=300$.

\subsection{The Template for Experiments}
\label{sec:template}

We adopt the following template for all experiments involving Qwen models, building the non-thinking template for Qwen3~\citep{yang2025qwen3}.

\begin{figure}[h]
\centering
\begin{tcolorbox}[
    colback=gray!5,
    colframe=black!70,
    width=\columnwidth,
    arc=3pt,
    boxrule=0.8pt,
    title={\textbf{Qwen3 non-thinking template}},
    fonttitle=\small
]
\small
\texttt{<|im\_start|>system}\\
Please reason step by step, and put your final answer within \texttt{\textbackslash boxed\{\}}.
\texttt{<|im\_end|>}

\vspace{0.3em}
\texttt{<|im\_start|>user}\\
\texttt{\{problem\}}\\
\texttt{<|im\_end|>}

\vspace{0.3em}
\texttt{<|im\_start|>assistant}\\
\texttt{<think>}\\
\\
\texttt{</think>}
\end{tcolorbox}
\caption{The Qwen3-NoThinking prompt template used for inference.}
\label{fig:qwen3_template}
\end{figure}

\subsection{Implementation of Reflection, Transition, and Execution
Thoughts}
\label{app:transition}

Following the reflection, transition, execution taxonomy defined in \citet{chen2025seal}, we classify each reasoning step into one of three categories using two types of heuristic rules. As shown in Table~\ref{tab:criteria}, the first is a \textit{prefix rule}: steps beginning with keywords such as 'wait' or 'alternatively' are classified as transition or reflection. The second is a \textit{phrase rule}: steps containing predefined mid-sentence phrases indicative of self-correction or redirection are likewise classified as transition or reflection. Steps satisfying neither criterion are classified as execution. Classification is case-insensitive. The proportion of each category is then computed over all steps in the generated responses.

\begin{table}[h]
    \centering
    \caption{Extraction of reflection, transition, and execution thoughts.}
    \label{tab:criteria}
    \begin{tabular}{m{2cm} m{1.5cm} m{8cm}}
        \toprule
        \textbf{Category} & \textbf{Type} & \textbf{Keywords / Phrases} \\
        \midrule
        \multirow{2}{*}{\textbf{Transition}} 
        & Prefix  & \textit{Alternatively} \\
        & Phrase  & \textit{think differently, another way, another approach, another method, another solution, another strategy, another technique} \\
        \midrule
        \multirow{2}{*}{\textbf{Reflection}}  
        & Prefix  & \textit{Wait} \\
        & Phrase  & \textit{verify, make sure, hold on, think again, 's correct, 's incorrect, Let me check, seems right} \\
        \bottomrule
    \end{tabular}
\end{table}

\subsection{Implementation of Reasoning Behaviors}
\label{app:behavior}
To analyze the evolution of reasoning behaviors during training, we follow \citet{zeng2025simplerl} and use GPT-4o~\citep{hurst2024gpt} to identify four key cognitive behaviors in model responses, with the annotation prompt provided in Figure~\ref{fig:behavior_prompt}. We report the proportion of responses exhibiting each behavior throughout training.
\begin{itemize}
\item \textbf{Backtracking}: The model actively identifies errors during response generation and explicitly revises previously used methods.
\item \textbf{Verification}: The model systematically checks intermediate results to ensure correctness.
\item \textbf{Subgoal Setting}: The model decomposes complex problems into smaller, manageable steps.
\item \textbf{Enumeration}: The model exhaustively considers multiple cases or possibilities to solve problems.
\end{itemize}

\begin{figure}[h]
\centering
\begin{tcolorbox}[
    colback=gray!5,
    colframe=black!70,
    width=\columnwidth,
    arc=3pt,
    boxrule=0.8pt,
    title={\textbf{Reasoning Evaluation Prompt}},
    fonttitle=\small
]
\small
Below is a chain-of-reasoning generated by a Language Model when attempting to solve a math problem. Evaluate this chain-of-reasoning to determine whether it demonstrates beneficial problem-solving behaviors that deviate from typical linear, monotonic reasoning patterns.

\vspace{0.5em}
\texttt{<start\_of\_reasoning>}\\
\texttt{\{segment\_text\}}\\
\texttt{<end\_of\_reasoning>}

\vspace{0.5em}
Specifically, identify and emphasize beneficial behaviors such as:

\begin{itemize}
    \item \textbf{Backtracking}: Explicitly revising approaches upon identifying errors or dead ends.
    \item \textbf{Verification}: Systematically checking intermediate results or reasoning steps.
    \item \textbf{Subgoal Setting}: Breaking down complex problems into smaller, manageable steps.
    \item \textbf{Enumeration}: Exhaustively considering multiple cases or possibilities.
\end{itemize}

\vspace{0.3em}
Additionally, remain attentive to other beneficial behaviors such as creative analogies, abstraction to simpler cases, or insightful generalizations. If no beneficial behaviors are observed, explicitly return \texttt{`none'}.

\vspace{0.5em}
\begin{tcolorbox}[
    colback=white,
    colframe=black!40,
    arc=2pt,
    boxrule=0.5pt,
    left=4pt, right=4pt, top=3pt, bottom=3pt
]
\textbf{Output format:}
\begin{verbatim}
{
  "behaviour": "",
  "example": ""
}
\end{verbatim}
\end{tcolorbox}

\end{tcolorbox}
\caption{System prompt used to evaluate beneficial reasoning behaviors in model-generated chain-of-thought, targeting non-linear problem-solving patterns including backtracking, verification, subgoal setting, and enumeration.}
\label{fig:behavior_prompt}
\end{figure}

 \begin{figure}[!t]
    \centering
    \includegraphics[width=1\linewidth]{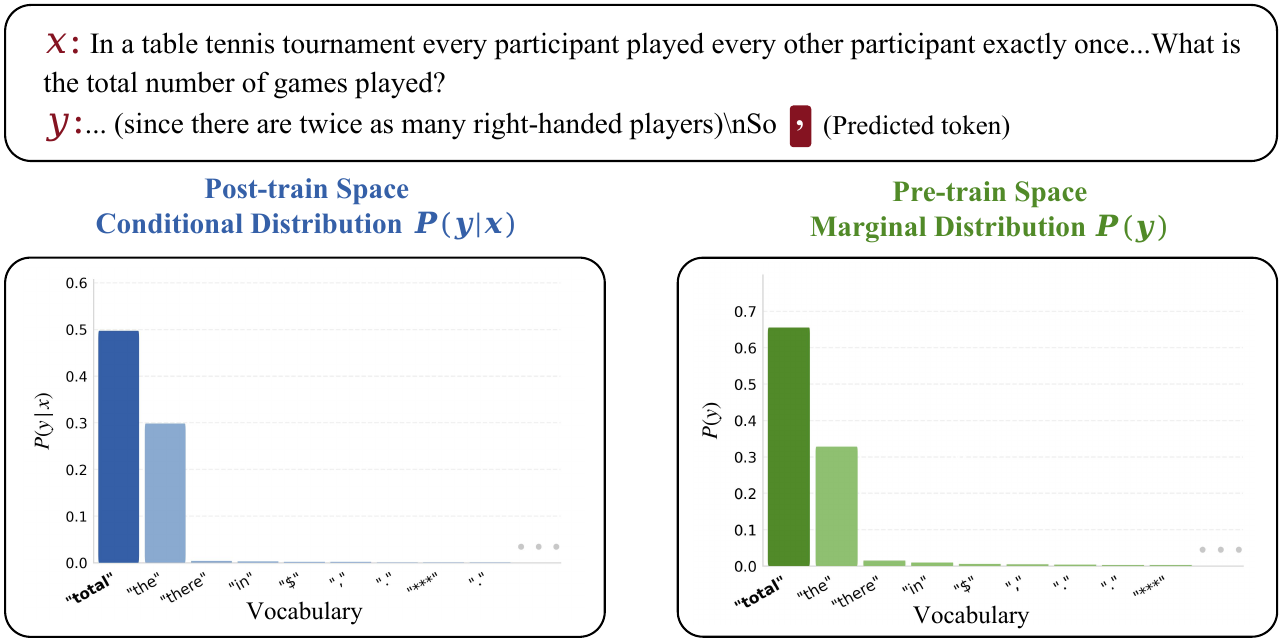}
    
    \caption{\textbf{Aligned case:} Token probability distribution with and without input conditioning under the same context. $x$ is a math problem and $y$ is a partially generated answer. The two distributions $P(y|x)$ and $P(y)$ show similar token rankings, suggesting the post-train space conditional distribution largely inherits the structure of the pre-train space marginal distribution.}
    \label{fig:token_distribution}
\end{figure}

\begin{figure}[!h]
    \centering
    \includegraphics[width=1\linewidth]{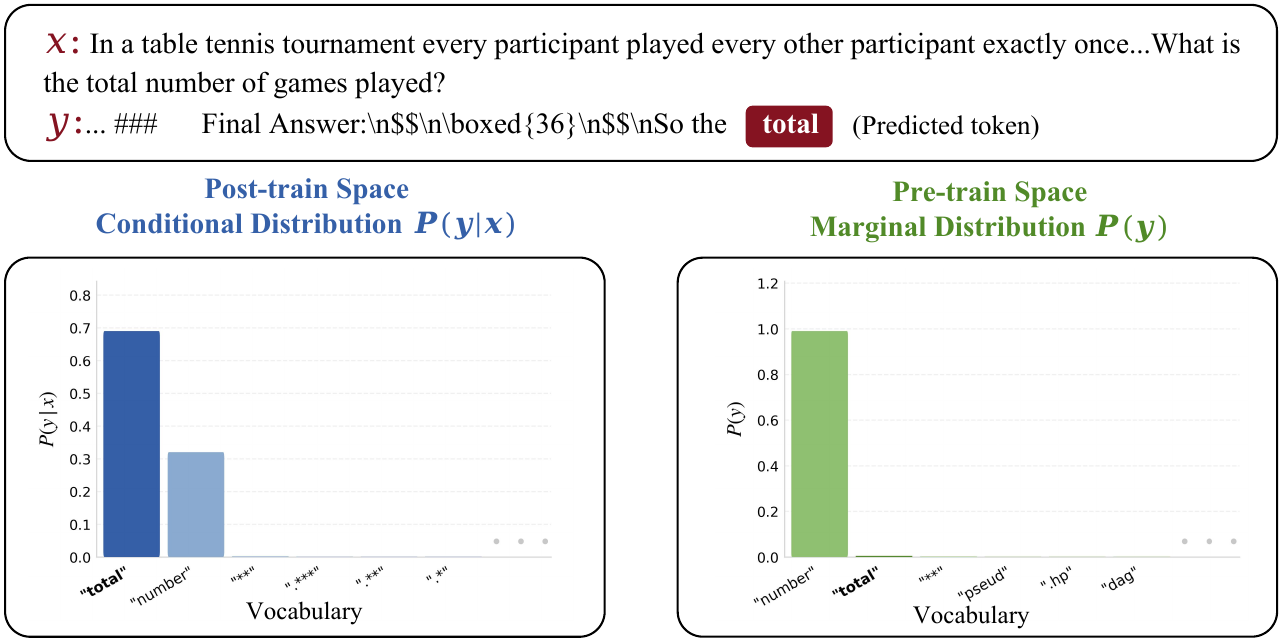}
    
    \caption{\textbf{Misaligned case:} Token probability distribution with and without input conditioning under the same context. $x$ is a math problem and $y$ is a partially generated answer. The top-1 token "total" under $P(y|x)$  receives near-zero probability under $P(y)$, indicating a significant discrepancy between the two distributions.}
    \label{fig:token_distribution_app}
\end{figure}

\begin{figure}[!t]
    \centering
    \captionsetup[subfigure]{justification=centering, skip=0pt}
    \begin{subfigure}{0.9\textwidth}
        \centering
        \includegraphics[width=0.9\linewidth]{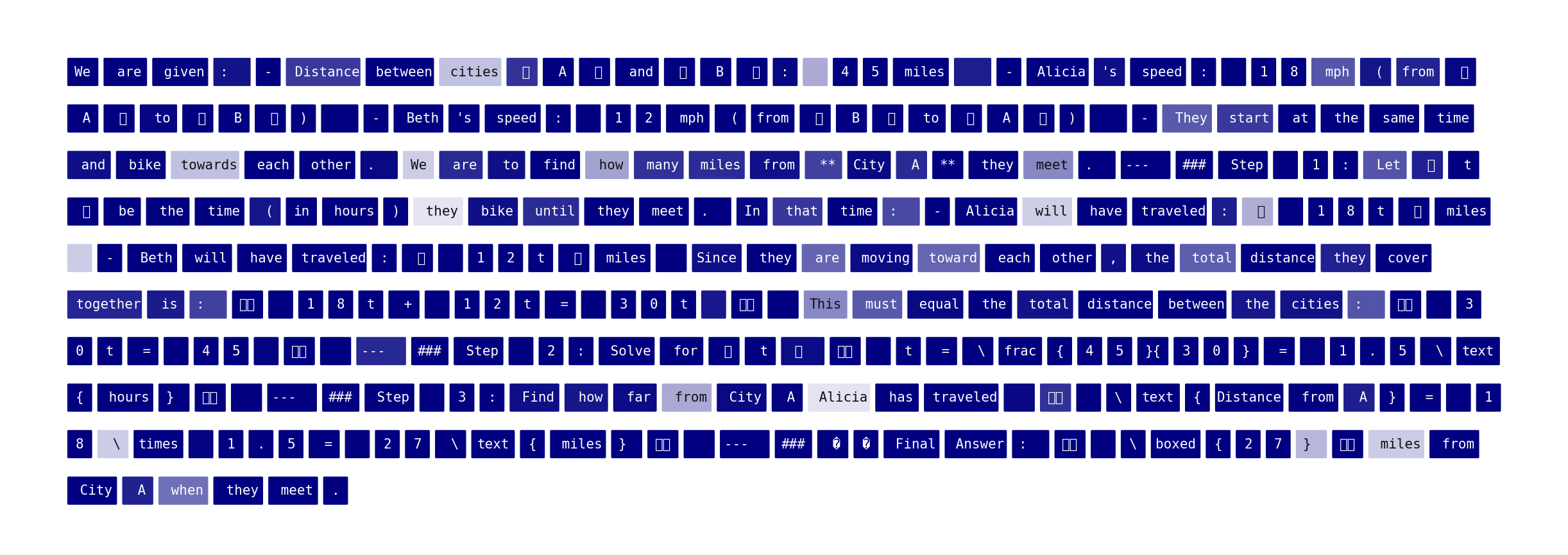}
        \caption{Conditional token log-probabilities: $\log P(y|x)$.}
        \label{fig:prob_y_given_x}
    \end{subfigure}
    \begin{subfigure}{0.9\textwidth}
        \centering
        \includegraphics[width=0.9\linewidth]{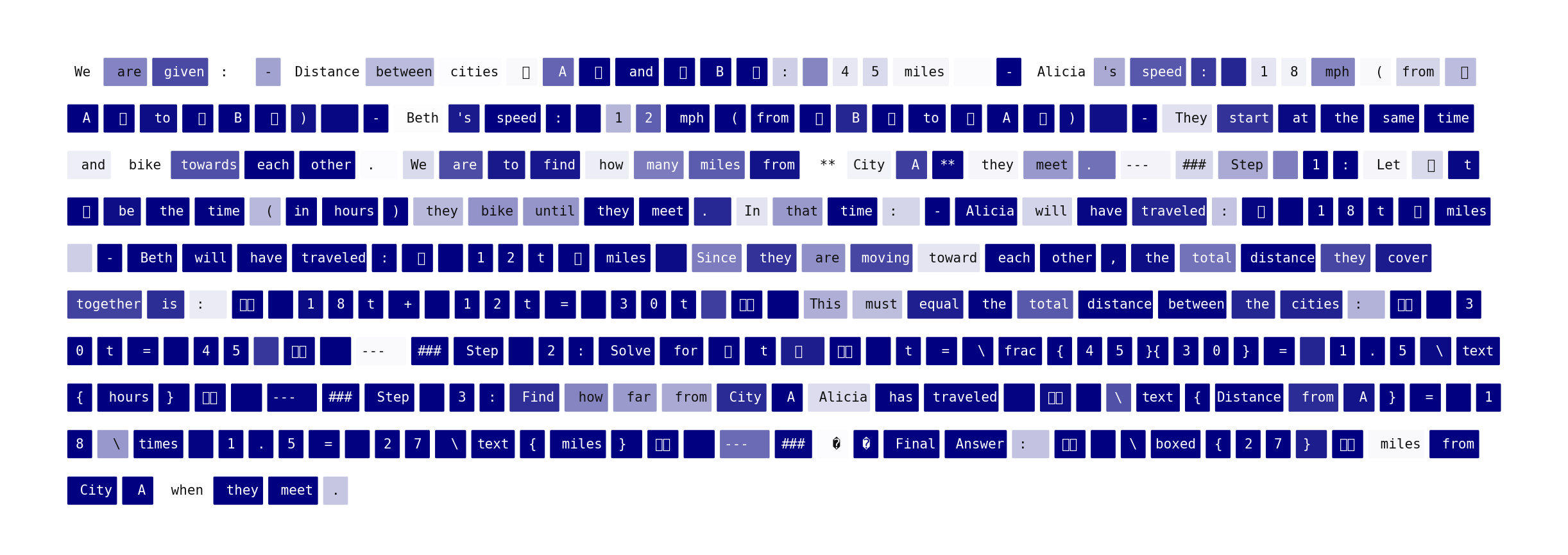}
        \caption{Marginal token log-probabilities: $\log P(y)$.}
        \label{fig:prob_y}
    \end{subfigure}
    \caption{Comparison of token-level generation log-probabilities with and without input conditioning.}
    \label{fig:prob_comparison}
\end{figure}

\begin{figure}[!h]
    \centering
    \includegraphics[width=1\linewidth]{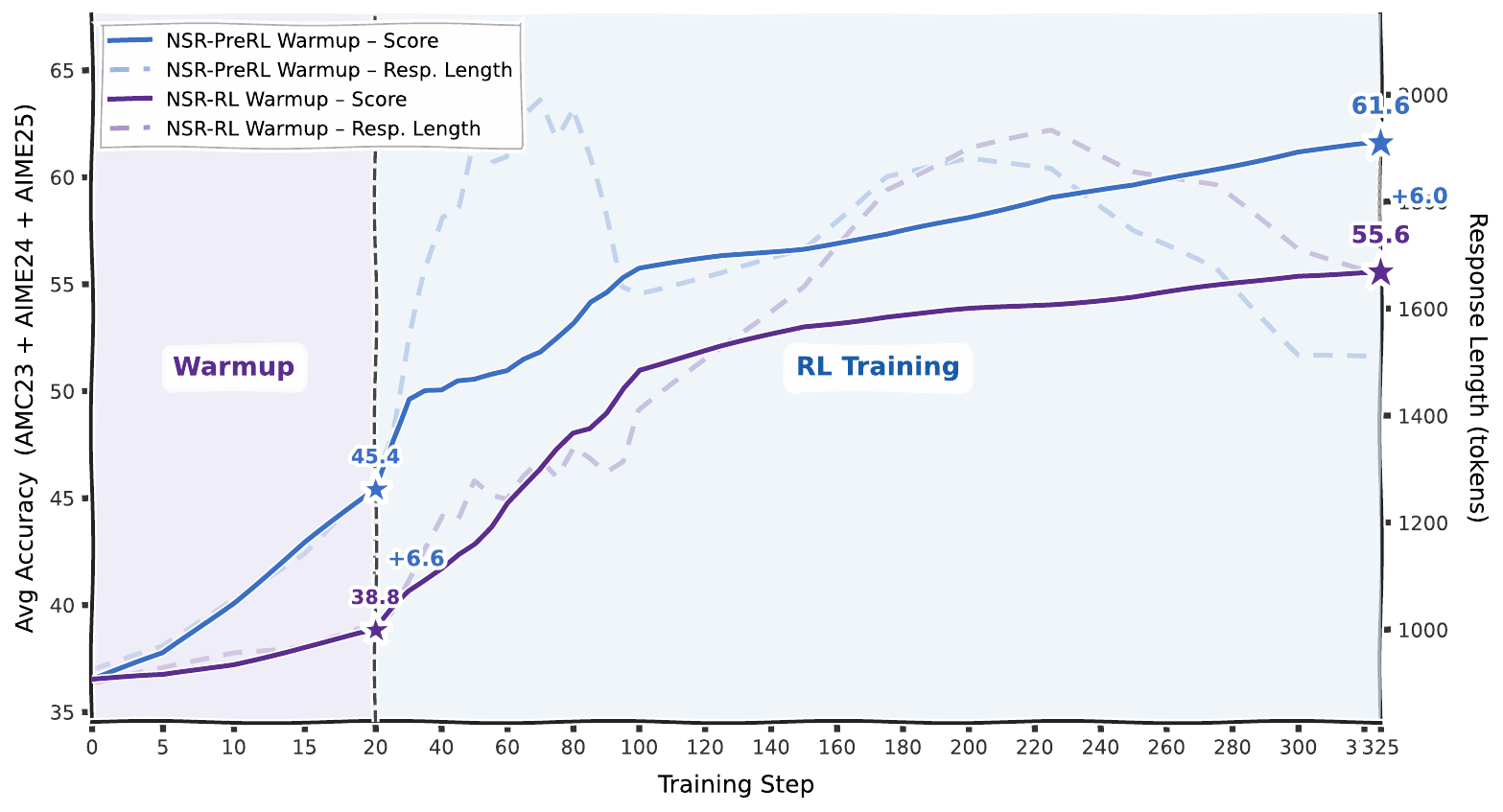}
    
    \caption{Training dynamics of NSR-PreRL warmup vs. NSR-RL warmup.}
    \label{fig:ablation_nsr_dynamic}
\end{figure}

\end{document}